# Make Your Bone Great Again : A study on Osteoporosis Classification

Rahul Paul, Saeed Alahamri, Sulav Malla, Ghulam Jilani Quadri

Department of Computer Science and Engineering
University of South Florida
Tampa, USA
{rahulp, saeed3, sulavmalla, ghulamjilani}@mail.usf.edu

ABSTRACT

Osteoporosis can be identified by looking at 2D x-ray images of the bone. The high degree of similarity between images of a healthy bone and a diseased one makes classification a challenge. A good bone texture characterization technique is essential for identifying osteoporosis cases. Standard texture feature extraction techniques like Local Binary Pattern (LBP), Gray Level Co-occurrence Matrix (GLCM), Law's etc. have been used for this purpose. In this paper, we draw a comparison between deep features extracted from convolution neural network against these traditional features. Our results show that deep features have more discriminative power as classifiers trained on them always outperform the ones trained on traditional features.

## I. INTRODUCTION

Osteoporosis is a disorder caused by lower mineral density, happens when there is disparity between growth and resorption of bones and increases the possibility of bone fractures. Osteoporosis comes from "osteo" or bones and "porosis" or porous which may lead to fragile and brittle bones which may break from falling or some minor injury, and become one of the major health problem in elderly people (>50 years), causing steep rise in healthcare costs [1]. Osteoporosis doesn't have any clear symptoms, so the person who is suffering from, may not know until he fractures a bone. Early screenings and diagnosis would require to effectively prevent osteoporosis [2]. The most commonly used method for diagnosis is dual energy x-ray absorptiometry (DEXA), which measure bone mineral density (BMD). Some other diagnostic imaging includes X-Ray, CT (Computed Tomography) scans, ultrasounds etc. Lately texture analysis [3-4] of bone structures using x-ray images shows simple way to analyze. But bone structures analysis based on texture is quite challenging as the osteoporotic or the control (healthy) cases both shows similar visual patterns.

Pothuaud [5] proposed trabecular bone texture analysis using the fractal analysis. Houam [6] presented an approach based on 1D projection on gray scale textures and lbp. 1D projection reduces redundant information and enhances shapes. Lbp was applied on the 1D signals obtained from different orientations for extracting features. Suprijanto [7] used gray level co-occurance matrix to extract features and support vector machine for classification. Song et al. [8] analyzed bone texture images using fisher encoding. Tafraouti [9] used fractal analysis and wavelet decomposition to classify osteoporotic and control patients. Their method was based on wavelet decomposition of the x-ray images into sub-bands and then extract features using the fractional Brownian motion analysis. Viet Quoc Ngo [10] proposed a combination approach based on Gabor filter, co-occurrence matrix and contourlet transform.

Recently in machine learning field, classification, object detection using convolutional neural network (CNN) has shown good performance. Artificial neural networks (ANN), which are inspired by human brain, have been used for classification and prediction. ANNs has mainly three layers input, output and hidden with activation function on hidden and output layers and layers consists of a number of interconnected nodes. As each consecutive layer has interconnection, the number of weights in between will increase rapidly, which is an issue with the ANNs. Convolutional neural network(CNN), a translational invariant neural network which consists of some convolutional layer and fully connected layers (same as ANNs). CNN uses several small filters on the input and subsampling the space of filter activations until there are sufficiently high level features.

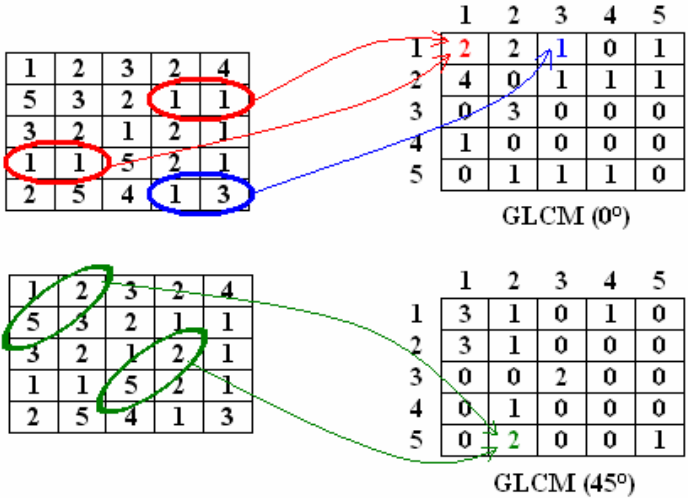

Figure 1. GLCM approach

Fukushima [11] proposed "neocognitron", a multilayered ANN, which has been used for recognition task of handwritten character. Later, a major advancement happened when, Yann LeCun [12] used deep convolutional network (LeNet) for handwritten zip code recognition. In the image classification field, different design of this basic architecture has been proposed and produced best results on ImageNet, MNIST, CIFAR datasets. The training of a CNN required sufficient amount of data as it has to learn millions of weights. Due to the less amount of data availability specially in the medical field, using a pre-trained CNN is useful. Donahue [13] examined, whether the features extracted from the activation of pre-trained CNN is useful for classification task. Pre-trained CNN means, training the CNN using a large dataset and utilize that previously learned knowledge to perform a new task, this is commonly called as transfer learning [14].

In this paper, we used pre-trained CNN models which is trained on ImageNet [15] to extract features from the x-ray images. We also implemented some traditional feature extraction methods such as run length matrix, GLCM (Gray level co-occurrence matrix) features and LBP (Local binary pattern). In this paper, we also experimented mixing deep features with traditional features.

The remainder of this paper is organized as follows: Section II and III explains traditional feature extraction approaches and transfer learning and pre-trained CNN architectures that we used for our study. Section IV describes classifiers used and feature selection methods. Section V analyzes the result using different approach. Section VI contains the conclusion.

## II. TRADITIONAL FEATURE APPROACHES

To get the texture of the given images (train set) we have use the traditional features extraction techniques such as GLCM (Gray level Co-occurrence matrix), LBP (local binary pattern), and GLRLM (Gray level run length matrix). GLCM is a method of texture features extractions. It basically gives a statistical measure of the texture of an image. The way it works is simply by observing how often a specific pattern occurs between two neighboring pixels. Some of the statistical measures returned by GLCM are Contrast, Energy, Correlations, Homogeneity, etc. The number of features we got from GLCM are 44 features for every image. Figure 1 shows GLCM for angle 0 and angle 45 [16].

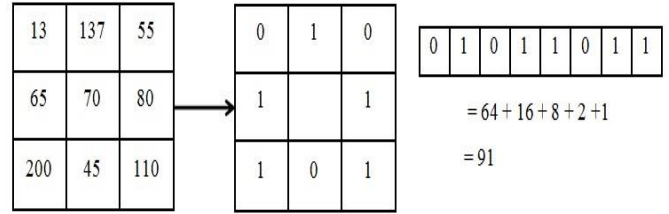

Figure 2. LBP

LBP is a descriptor for texture Spectrum model in Computer Vision. LBP examine the 8 neighbors of each pixel. 8 neighbors are compared with their center value. If the center value is greater than a neighbor, then place zero (0) at that neighbor. Otherwise, place one (1). All neighbors are examined the same way for every pixel in the image. The return image is a binary image with (zeros and ones). For this problem, we used the basic LBP which compute the binary image and return the LBP feature as one number of each image. The following example shows how the LBP works [17].

GLRLM captures the texture pattern within a specific sequence and direction; whereas, GLCM is concern about just pairs of pixels. The matrix generated by GLRLM has the number of columns equal to the number of intensity of the gray image, and the rows equal to the length of runs sequence [18]. After the GLRLM has been generated the features created from GLRLM are 5 features including run length emphasis, run length non-uniformity, etc.

## III. CONVOLUTIONAL NEURAL NETWORK AND TRANSFER LEARNING

A Convolutional Neural Network [19] is biologically influenced variants of multi-layer feed forward model that are currently widely using in image classification and recognition tasks. The layers of a CNN are mainly classified into four layers or operations. First layer is convolutional layer, which is based on a mathematical approach called convolution, which takes a kernel and applied it to all over an input image to generate a filtered image. Second operation is Pooling or sub-sampling which reduces the dimensionality of the feature map although keep most important information for next convolution layer. Third layer is activation function layer, which applies activation function elementwise and got the output by checking whether the neurons are fired or not. Some of the common activations functions are rectified linear unit or ReLU, sigmoid functions, tanh functions etc. The last layer is fully connected layer, where all the neurons from the previous layer is connected to all nodes in the adjacent layers which is same as the fully connected multilayer perceptron. CNN must learn lots of weights and for that we need lots of data for training, but in medical data doesn't have large enough data to be used for training from scratch. To counter this problem for training of CNN an alternative approach Transfer learning [20-21] can be used. Transfer learning is an approach where knowledge leaned from previous task can be applied to some new task domain. In our study, we have used transfer learning concept, by using pre-trained CNN which is trained on ImageNet [15] dataset. ImageNet is one of the most popular image database, consists of more than 14 million images of 1000 distinct object class.

| Architecture | CNN-F |
|---|---|
| conv 1 | 64 x 11 x 11 st. 4, pad 0 |
| conv 2 | 256 x 5 x 5 st. 1, pad 2 |
| conv 3 | 256 x 3 x 3 st. 1, pad 1 |
| conv 4 | 256 x 3 x 3 st. 1, pad 1 |
| conv 5 | 256 x 3 x 3 st. 1, pad 1 |
| full 6 | 4096 dropout |
| full 7 | 4096 dropout |
| full 8 | 1000softmax |

Table 1. Vgg-f architecture

| Architecture | CNN-M |
|---|---|
| conv 1 | 96 x 7 x 7 st. 2, pad 0 |
| conv 2 | 256 x 5 x 5 st. 2, pad 1 |
| conv 3 | 512 x 3 x 3 st. 1, pad 1 |
| conv 4 | 512 x 3 x 3 st. 1, pad 1 |
| conv 5 | 512 x 3 x 3 st. 1, pad 1 |
| full 6 | 4096 dropout |
| full 7 | 4096 dropout |
| full 8 | 1000softmax |

Table 2. Vgg-m architecture

| Architecture | CNN-S |
|---|---|
| conv 1 | 96 x 7 x 7 st. 2, pad 0 |
| conv 2 | 256 x 5 x 5 st. 1, pad 1 |
| conv 3 | 512 x 3 x 3 st. 1, pad 1 |
| conv 4 | 512 x 3 x 3 st. 1, pad 1 |
| conv 5 | 512 x 3 x 3 st. 1, pad 1 |
| full 6 | 4096 dropout |
| full 7 | 4096 dropout |
| full 8 | 1000softmax |

Table 3. vgg-s architecture

We have used three different pre-trained CNN architectures (vgg-m, vgg-f, vgg-s) as described in Chatfield's [22] work. The details of these architecture are in the Table 1 to 3. Each CNN has five convolution layer and followed by three fully connected layer. We are using pre-trained network to extract features from last hidden layer after applying the activation function (post relu) [23]. In this experiment, x-ray images are used which are different than the images in ImageNet database, but we are hypothesizing some useful texture features might exist. The pre-trained CNN that we used here is implemented in a matlab called matconvnet [24]. The input image size is 224x224 for the CNN architectures so we use bi-cubic interpolation to resize the input images. The x-ray images are grayscale, so we modified the code and extracting features using only R channel and ignoring B and G channel. The deep features that we are extracting is of 4096 dimension.

IV. FEATURE SELECTORS AND CLASSIFIERS

Here we used two feature selectors and five different classifiers.

*Symmetric Uncertainty*

Symmetric Uncertainty [25] is a correlation based filter approach used for feature selection. It ranks the features in the data according to its relevance to the class while at the same time are not redundant when considered with other relevant features. Symmetric Uncertainty (SU) between an attribute and class is defined as follows

$$SU(Class, Attribute) = 2[\frac{H(Class)-H(Class|Attribute)}{H(Class) + H(Attribute)}]$$

where $H$ is the entropy, a measure of uncertainty, of random variable. If $P(x_i)$ is the probability of all values of a random variable $X$, then the entropy would be

$$H(X) = -\sum_i P(x_i)log_2(P(x_i))$$

*Relief-F*

Relief-F [26], [27] is a noise and missing data tolerant, multi-class feature selection algorithm based on instance-based learning approach. It searches for nearest neighbors, ones from the same class (nearest hits) and others from different class (nearest misses), of an instance to calculate weights for each attribute. The logic behind this is that a good attribute, which must separate instances belonging to different class and have same value for ones belonging to the same class, gets more weight.

*Naïve Bayes*

Naïve Bayes [28] is a simple probabilistic classification algorithm that is based on the Bayes' theorem of conditional probability. Given an evidence $E$ for a hypothesis $H$, Bayes' theorem calculates the probability of $H$ given $E$ as follows.

$$P(H|E) = \frac{P(E|H).P(H)}{P(E)}$$

Hypothesis can be an instance belonging to a class and evidence can be the attribute values of that instance. Hence, we can calculate the probability $P(H|E)$ for the instance belonging to each class and assign the class label for which this probability is highest. For a given the class label, we assume that the features are independent of each other, which is a simplistic one. But this technique seems to perform well on practice and requires only a small amount of training data.

*Support Vector Machine*

Support vector machine (SVM) [29] is a technique of learning a maximum-margin hyperplane that can separate instances of different classes. This maximum-margin hyperplane is the one that has the greatest separation between the instances of the different class and can be learned using the method of least squares. SVM avoids overfitting as the maximum-margin hyperplane is relatively stable as it depends on only those instances that are closest to it, also called support vectors. Moreover, non-linear classification is possible by applying various kernel functions (polynomial, sigmoidal, radial basis function) such that the instance space is transformed to a higher dimensional space where they may be linearly separable.

*Decision Trees*

Decision tree [30] is a divide-and-conquer strategy that splits the given instances recursively such that instances belonging to the same class end up together. The process of recursive splitting forms a tree with nodes and leaves. The nodes contain test condition on attribute values such that only a subset of examples pass down each branch. Ultimately, we end up with leaves that contain pure or nearly pure examples (belonging to same class). This class label of each leaf is the prediction by the decision tree. An attribute is selected as the test attribute at a node if it has the highest information gain.

*Random Forest*

Random forest [31] is a classification technique that trains multiple decision trees on different subset of features of training instances. While a single decision tree is easy to create and fast during prediction, it can overfit the training data. Hence, an ensemble of decision trees (a forest) is created such that each one trains on randomly selected subset of features and during prediction of test set, we choose the class which receives the majority vote by these trees. This way, a general concept is learned (avoiding overfitting) without decreasing the accuracy on the training set.

*Nearest Neighbor*

Nearest neighbor [32] or instance based classification is a lazy classification technique in which we simply store all the training set. During testing, we assign the test instance to the class of its nearest neighbors. Distance between a test instance to those of training can be measured using various distance measure (Euclidean, Manhattan etc.). While training is fast, it may be memory intensive as we need to store all the instances as it is. Furthermore, testing can be slow as we need to calculate the distance of a test instance with all the training instances and find the minimum.

V. EXPERIMENT

*Dataset*

The dataset used consisted of 174 bone x-ray images obtained from IEEE-ISBI 2014 competition dataset ( http://www.univ-orleans.fr/i3mto/challenge-ieee-isbi-bone-texture-characterization) . Out of which 58 images are using for testing and doesn't have any labels or class given and 116 images are equally subdivided into control and osteoporosis cases.

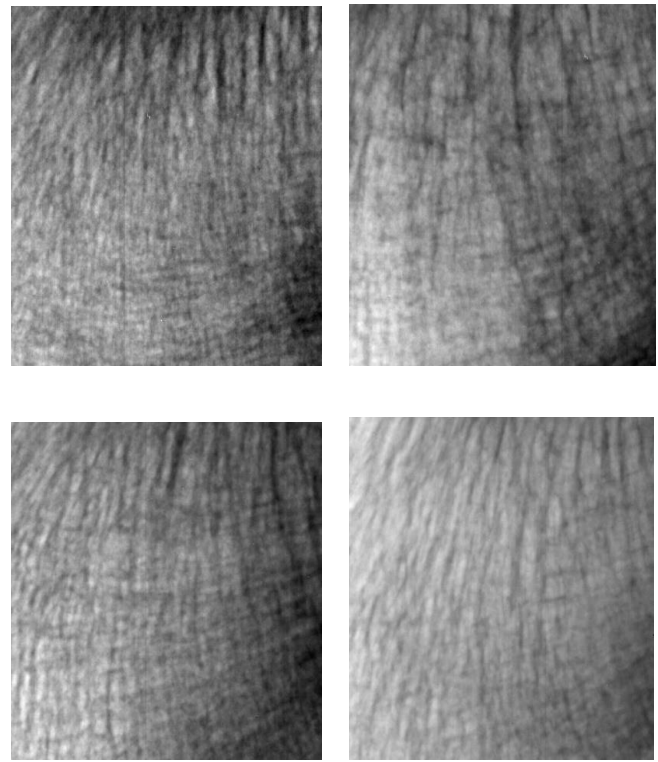

Figure 3. Example of the x-ray Osteoporotic (top row) and control subjects (bottom row)

*Approach 1: pre-trained CNN*

The pre-trained CNN used in our study required 224x224 input image, so by using bi-cubic interpolation method we resized the images. The size of extracted deep features vector from each x-ray images were 4096. Classification using these many features might not be useful, so we used some feature selection approaches. In our study, we used symmetric uncertainty and relief-f for feature selection on the training dataset using 10fold cross validation. We selected 10,15 and 20 features per fold for our selected classifiers e.g. Random forests, decision tree, naïve Bayes, SVM and Nearest neighbor classifiers.

The best result of 70.6897% (AUC – 0.742) was obtained using post-relu features from a vgg-f CNN architecture and a nearest neighbor classifier in a 10-fold cross validation with ten features using the relief-f feature ranking algorithm on each fold.

With vgg-m CNN architecture the best result obtained is 79.3103% (AUC- 0.85) using a random forests classifier with fifteen features obtained using symmetric uncertainty feature ranking algorithm in a 10-fold cross validation. Using vgg-s pre-trained architecture the best result of 76.7241%(AUC-0.813) obtained using a random forests classifier in a 10-fold cross validation with twenty features obtained using symmetric uncertainty feature ranking algorithm.

*Approach 2: Traditional approach*

We used GLCM, run length and LBP features for classification on the training set using 10-fold cross validation. We used all the features, as well as top 10,15,20 features selected by symmetric uncertainty, relief-f and t-test for classification.

Using all traditional features, best accuracy was 60.3448% (AUC-0.603) using a SVM classifier. The best result of 57.7586% (AUC -0.58) using a nearest neighbor classifier with ten features obtained using relief-f feature ranking algorithm in a 10-fold cross validation. Using fifteen features, the best result of 56.8966% (AUC-0.569) was obtained using SVM classifier and symmetric uncertainty feature ranking algorithm in a 10-fold cross validation. With twenty features, best accuracy of 59.4828%(AUC-0.597) was obtained using a random forests classifier and relief-f feature ranking algorithm in a 10-fold cross validation.

*Approach 3. Merging traditional and deep feature*

In this new approach we merged top ten, fifteen and twenty deep features with the traditional features. From both deep feature and traditional feature vector, top ten, fifteen and twenty features were selected per fold using symmetric uncertainty, relief-f and t-test feature ranking algorithm and merged them together to make twenty, thirty and forty features.

The best accuracy of 75.8621%(AUC-0.789) was obtained using a random forests classifier with symmetric uncertainty feature selector from twenty merged features.
Using thirty merged features, the best accuracy obtained was 74.1379%(AUC-0.772) using a random forests classifier with symmetric uncertainty feature selector. With forty merged features, best accuracy of 72.4138%(0.766) was obtained using a random forests classifier with symmetric uncertainty feature selector.

Table 4 summarizes the best results obtained using with traditional feature and deep features alone and with merged features. Table 5 explains some statistical analysis between the best results of different approaches.

| Feature type | Deep features (vgg-m) | Traditional Feature | Traditional Feature | Mixed (Deep+ traditional quantitative) features |
|---|---|---|---|---|
| Classifier used | Random Forests | Random Forest | SVM | Random Forests |
| Feature selector used | Symmetric uncertainity | Relief-f | None | Symmetric uncertainity |
| Number of features | 15 | 20 | ALL | 20 (10 deep+ 10 traditional features) |
| Accuracy | **79.3103** | 59.4828% | 60.3448 | 75.8621% |
| AUC | **0.85** | 0.597 | 0.603 | 0.789 |

Table 4. Selected Results

We choose three sets of feature one from each approach traditional feature and deep features alone and with merged features for evaluating the results on the test images. Table 6 summarizes the results on the test data.

| Feature type | Deep features & Traditional Feature | Traditional Feature & Mixed Feature |
|---|---|---|
| P value (two-tailed) | 0.000075 The result is significant at $p < 0.05$. | 0.004428 The result is significant at $p < 0.05$. |
| z-value | 3.9615 | -2.8459 |

Table 5. Statistical Analysis

| Feature type | Deep features |
|---|---|
| Classifier used | Random Forests |
| Feature selector used | Symmetric uncertainty |
| Number of features | 15 |
| Accuracy | **44.82%** (TP-12, FP-15, TN-14 FN-17) |
| Sensitivity | 0.414 |
| Specificity | 0.4827 |

Table 6. Results on Blind Data

## VI. CONCLUSION

Classify osteoporosis by considering x-ray images is very difficult as the x-ray images obtained from the healthy patient looks very similar to that of the osteoporotic patient. Various traditional features have already been used to classify the osteoporotic cases from the control cases. Recently convolutional neural network is using widely using for classification and extracting features. But training a convolutional neural network need huge amount of data. In this experiment, we only have very small number of training data, which is not sufficient to train a CNN. To solve this problem, we use transfer learning approach – pre-trained CNN. The pre-trained CNNs are trained on ImageNet data and we are using the pre-trained CNNs to extract features from the last hidden layer after applying the activation function. In our study we took three different approach: classification using traditional features (GLCM, LBP, RLM), classification using deep features and merging top deep and traditional features. On the training set, the best result of 79.3103% (0.85) was obtained from fifteen deep features selected by symmetric uncertainty feature ranking algorithm and with a random forests classifier, from the post-vggm pre-trained CNN and using. Using traditional feature only, best result of 60.3448 (0.603) was obtained from the training set using all the feature vectors. By merging the deep and traditional features, we got the best accuracy of 75.8621% (AUC-0.789) from the training set using a random forests classifier by merging ten traditional features and ten deep features obtained from the post-vggm pre-trained CNN and using symmetric uncertainty feature ranking algorithm. We now have these three sets of feature using which we can do our prediction on blind test set. Using only deep features, best result obtained is 44.82% with sensitivity 0.414 and specificity 0.4827. Our next work consists of working more on deep feature to generate a better classification result on blind test data and tuning the CNN.

ACKNOWLEDGMENT

The authors would like to thank Dr. Lawrence Hall for providing us this opportunity to work on a data mining project and his constant support throughout the process.